\pdfoutput=1

\documentclass[11pt]{article}

\usepackage[preprint]{acl}

\usepackage{times}
\usepackage{latexsym}

\usepackage[T1]{fontenc}

\usepackage[utf8]{inputenc}

\usepackage{microtype}

\usepackage{inconsolata}

\usepackage{graphicx}
\usepackage{algorithm}
\usepackage{algorithmic}
\usepackage{threeparttable}
\usepackage{multirow,booktabs,amssymb,amsmath}

\usepackage{hyperref}
\usepackage{colortbl}        
\usepackage{array}
\title{DynaSearcher: Dynamic Knowledge Graph Augmented Search Agent \\ via Multi-Reward Reinforcement Learning}

\author{Chuzhan Hao\quad
Wenfeng Feng\quad
Yuewei Zhang\quad
Hao Wang\protect \thanks{Corresponding author.} 
\\
Alibaba Cloud Computing
\\
\tt {haochuzhan.hcz@alibaba-inc.com, cashenry@126.com}
}

\begin{document}
\maketitle
\begin{abstract}
Multi-step agentic retrieval systems based on large language models (LLMs) have demonstrated remarkable performance in complex information search tasks. 
However, in practical applications, these systems are limited by factually inconsistent intermediate queries and inefficient search trajectories, which can cause reasoning deviations and redundant computations.
To address these issues, we propose DynaSearcher, an innovative search agent enhanced by dynamic knowledge graphs and multi-reward reinforcement learning (RL). Specifically, our system leverages knowledge graphs as external structured knowledge to guide the search process by explicitly modeling entity relationships, thereby ensuring factual consistency in intermediate queries and mitigating biases from irrelevant information. Furthermore, we employ a multi-reward RL framework for fine-grained control over training objectives such as retrieval accuracy, efficiency, and response quality. This framework promotes the generation of high-quality intermediate queries and comprehensive final answers, while discouraging unnecessary exploration and minimizing information omissions or redundancy. Experimental results demonstrate that our approach achieves state-of-the-art answer accuracy on six multi-hop question answering datasets and exhibits strong generalization and robustness across diverse retrieval environments and larger-scale models, highlighting its broad applicability\footnote{The model is available at \url{https://modelscope.cn/collections/DynaSearcher-a00139d1ef2542}.}.

\end{abstract}

\section{Introduction}
Large language models have demonstrated remarkable capabilities in task planning and agentic reasoning, with reinforcement learning significantly improving their performance on complex reasoning tasks~\cite{shao2024deepseekmath,guo2025deepseek}. However, reliance on static parametric knowledge presents notable limitations, often leading to hallucinations and inefficient reasoning. To tackle these challenges, it is crucial to explore how to efficiently access diverse external information to support LLMs in achieving deliberate and well-substantiated reasoning. Therefore, a novel search paradigm termed \textit{Agentic Deep Research Systems} has gradually become an important research task~\cite{li2025searcho1,jin2025searchr1,chen2025research,zhang2025websearch}.

Previous research has utilized Chain-of-Thought (CoT)~\cite{wei2022chainofthought} prompting to decompose complex problems into sequential sub-tasks, subsequently leveraging external information dynamically to bridge knowledge gaps and tackle intricate reasoning tasks~\cite{trivedi2022ircot,shao2023iterretgen,yue2024iterdrag,feng2025airrag}. However, these approaches remain highly sensitive to various prompt formulations. \citet{feng2025airrag} employs the Monte Carlo Tree Search (MCTS) to design more sophisticated reasoning frameworks, achieving deep exploration of the solution space. This process incurs significant inference overhead, which reduces the likelihood of its widespread practical application. Meanwhile, these prompt-based methods incorporate singular strategic planning and iterative refinement capabilities throughout the pipeline, failing to fully exploit the agentic potential of LLMs. Furthermore, \citet{li2025searcho1} integrates an agentic search workflow into the reasoning process, enabling dynamic retrieval when dealing with uncertain or incomplete information.

Recently, reinforcement learning has achieved remarkable success in mathematical reasoning and decision-making scenarios~\cite{guo2025deepseek}. \citet{jin2025searchr1,song2025r1searcher} also utilize RL to significantly enhance the capability of small language models (SLMs) to address complex multi-hop reasoning tasks. These training-based approaches directly equip LLMs with the ability to autonomously use external retrieval tools during the end-to-end training process, thereby achieving dynamic interaction with the external environment throughout training~\cite{zheng2025deepresearcher,chen2025research}. Owing to their superior agentic abilities, strong generalization, and efficient reasoning processes, RL-based agentic search approaches are increasingly emerging as a significant trend in deep research~\cite{zhang2025websearch}. 
However, current RL-based search agents rely on a single search tool and coarse global rewards, lacking effective guidance for intermediate query generation and struggling to explore efficient reasoning trajectories. The presence of substantial noise in unstructured text information can easily lead to deviations from the correct reasoning path or introduce redundant computations. In addition, both coarse-grained global rewards and existing step-wise rewards are typically simple aggregations of distinct reward signals, failing to fully leverage the interdependencies among different rewards to enhance explicit guidance on query trajectories and intermediate query generation.

To address these challenges, we propose DynaSearcher, a dynamic knowledge graph augmented multi-reward reinforcement learning framework tailored for search agents. 
Specifically, we employ knowledge graphs (KGs) as structured external knowledge, explicitly modeling entity relationships during multi-step reasoning to guide the search process towards factually aligned intermediate queries and reduce deviation caused by noise or irrelevant information.
Furthermore, we design a multi-reward mechanism that balances retrieval accuracy, efficiency, and the final response quality during the RL training process. This mechanism not only encourages high-quality intermediate query generation while penalizing excessive search steps to prevent unnecessary exploration or premature termination, but also incentivizes LLMs to produce comprehensive and accurate final responses, avoiding information omission or redundancy. Experimental results demonstrate that our method achieves state-of-the-art performance on various complex multi-hop question answering (QA) datasets, while maintaining comparable effectiveness under low-resource context length settings. We further validate the generalization and robustness of our approach across different retrieval environments and with larger-scale models, indicating its broad applicability. In summary, our main contributions are as follows:

\begin{itemize}
    \item We introduce knowledge graphs as an external structured knowledge source, guiding the search process through explicit modeling of entity relationships. This ensures the consistency of intermediate queries with factual information while mitigating deviations induced by noise or irrelevant information, thereby significantly enhancing the performance of the search agent.
    \item We propose multi-reward reinforcement learning that integrates both gain and penalty rewards into the conventional outcome-based reward framework. By incorporating fine-grained feedback, the approach provides nuanced guidance, fostering robustness and reliability in handling complex tasks.
    \item Extensive experiments demonstrate that DynaSearcher significantly outperforms existing search agents based on reinforcement learning. Our approach exhibits broad applicability across diverse retrieval environments and larger-scale models, while maintaining its performance even in low-resource settings.
\end{itemize}

\section{Related Work}
\subsection{Retrieval-Augmented Generation}
Early retrieval-augmented generation (RAG) approaches employ various strategies such as branching, iteration, and adaptive retrieval to solve complex tasks. These methods rely on manually crafted workflows to guide LLMs in interacting with external knowledge sources. IRCoT~\cite{trivedi2022ircot} leverages CoT to steer the retrieval process, refining CoT with the retrieved information. \citet{press2022selfask,asai2023selfrag,yue2024iterdrag} refine intermediate queries to acquire valuable knowledge through multi-turn iterations. AirRAG~\cite{feng2025airrag} applies MCTS to dynamically explore the reasoning paths. However, these approaches are limited to manually designed prompts and workflows, failing to fully unleash the inherent reasoning potential of LLMs.

\subsection{Autonomous Search Agents}
As the reasoning and decision-making capabilities of the foundation models continue to improve, Search-o1~\cite{li2025searcho1} significantly improves model performance in complex scenarios by designing an agentic search workflow, providing superior flexibility and generalization. DeepSeek-R1~\cite{guo2025deepseek} also demonstrates that outcome-based RL can significantly enhance the autonomous reasoning and decision-making capabilities of models. Therefore, RL has been applied to various complex reasoning tasks and agent-based scenarios. Complex multi-hop question answering represents a typical integrated application scenario that heavily relies on model-driven planning and reasoning. \citet{song2025r1searcher,chen2025research,jin2025searchr1} have successfully applied end-to-end reinforcement learning to complex agentic search scenarios, further advancing the development of agentic deep research systems. These methods autonomously select retrieval tools during the reasoning process to interact with external environments.
DeepResearcher~\cite{zheng2025deepresearcher} scales RL in real-world environments by incorporating authentic web search interactions. s3~\cite{jiang2025s3} decouples the searcher from the generator and trains the searcher with fewer samples. EvolveSearch~\cite{zhang2025evolvesearch} further explores the self-evolution process of search agents. StepSearch~\cite{wang2025stepsearch} introduces fine-grained reward signals to steer strategic query planning and improve retrieval quality in complex search environments.

In contrast to such methods, our proposed DynaSearcher incorporates structured knowledge graph information to dynamically model entity relationships during multi-step reasoning, effectively reducing reasoning path deviations and redundant computations caused by irrelevant information. Furthermore, we design the multi-reward reinforcement learning by leveraging the dependencies among rewards at different granularities, introducing gain and penalty rewards to provide precise guidance beyond the outcome-based reward. In the experiment, we thoroughly validate the performance gains and efficient reasoning paths achieved through the integration of structured knowledge graph and the multi-reward reinforcement learning.

\begin{figure*}[t]
\centering
\includegraphics[width=1.00\textwidth]{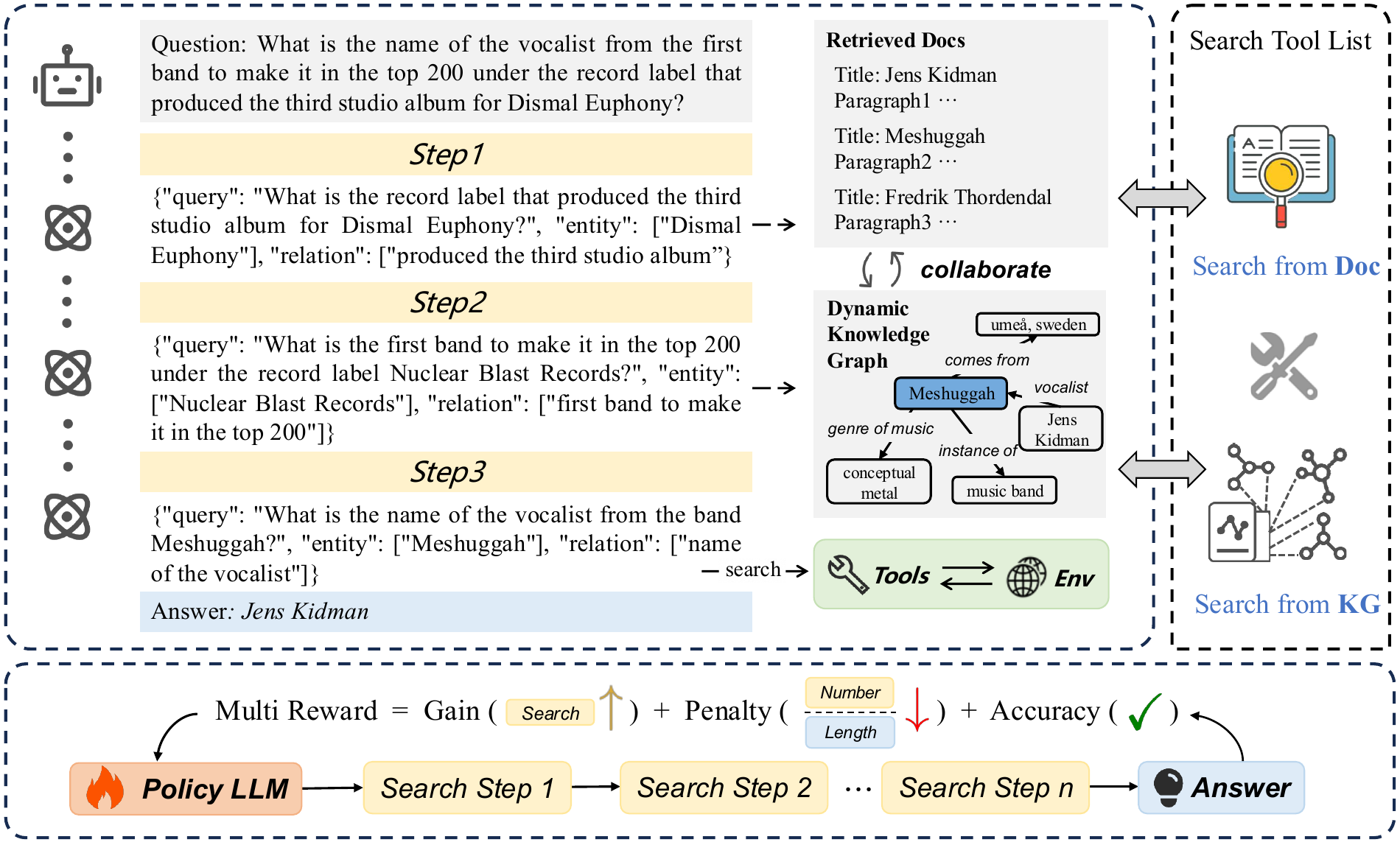} 
\caption{The framework of our proposed DynaSearcher.}
\label{framework}
\end{figure*}

\section{Methodology}
In this section, we introduce DynaSearcher, an advanced and efficient search agent that leverages dynamic knowledge graphs and multi-reward  reinforcement learning to guide the generation of intermediate queries during multi-step reasoning. It effectively mitigates reasoning deviations induced by irrelevant information, enabling efficient retrieval processes and precise answer generation. The overview of DynaSearcher is illustrated in Figure~\ref{framework}.

\subsection{Rollout with Search Tools} \label{sec:rollout}
Inspired by Search-o1~\cite{li2025searcho1}, current advanced RAG methods introduce an agentic search strategy, transforming the exploration process into an iterative interaction between the intrinsic reasoning of LLMs and the external environment, thus effectively activating their autonomous reasoning capabilities. During interactions with the external environment, these methods often rely on unstructured text retrieval systems to supplement information for intermediate reasoning steps. Irrelevant textual noise can easily result in inefficient intermediate queries. Therefore, we dynamically retrieve relevant knowledge graphs $\mathcal{G}$ during the reasoning process to guide the generation of intermediate queries.

\noindent\textbf{Doc Search Tool}. We employ two types of document search tools: a locally deployed vector-based retrieval service and the web search provided by Tavily\footnote{\url{https://tavily.com}.}. Both tools return text-based retrieval results. In the experiments, we fix a top-k value (e.g., 5) to control the number of retrieved documents.

\noindent\textbf{KG Search Tool}. We utilize Wikidata5M~\cite{wang2021wikidata} as the source for our structured knowledge graph, which provides a more precise representation of semantic relationships. During intermediate reasoning steps, we retrieve relevant single-hop knowledge subgraphs based on the parsed entities, supporting subsequent reasoning processes.

As illustrated in Figure~\ref{framework}, our approach integrates both unstructured document and structured KG retrieval tools, enabling queries across different knowledge sources. In each reasoning trajectory, DynaSearcher begins with reflective analysis and strategic planning based on the input question, then generates a JSON-formatted request specifying subqueries as well as extracted relevant entities and relationships. It then invokes both search tools to interact with the external environment and gather information. Specifically, we adopt an iterative reasoning-retrieval loop, similar to~\citet{chen2025research} and detailed in Table~\ref{system_template}, where reasoning and retrieval alternate. By collaboratively integrating these tools, DynaSearcher is able to explore more efficient and effective reasoning paths.

\begin{table}[ht]
\centering
    \renewcommand{\arraystretch}{0.8}
\small
\begin{tabular}{p{7.2cm}}
\toprule
\toprule
You are a helpful assistant that can solve the given question step by step with the help of the wikipedia search tool. \
Given a question, you need to first think about the reasoning process in the mind and then provide the answer. \
During thinking, you can invoke the wikipedia search tool to search for fact information about specific topics if needed. \
The reasoning process and answer are enclosed within \texttt{<think>} \texttt{</think>} and \texttt{<answer>} \texttt{</answer>} tags respectively, \
and the search input and result are enclosed within \texttt{<search>} \texttt{</search>} and \texttt{<result> </result>} tags respectively. \
Search input is json format like \{``query'': ``xxx'', ``entity'': [``yyy''], ``relation'': [``zzz'']\} and applied to the search tools, \
where query is used to search wikipedia articles, entity(s) and relation(s) are used to search wikidata, a knowledge base of entities and relations. \

For example, \texttt{<think>} This is the reasoning process. \texttt{</think>} \texttt{<search>} \{``query'': ``Who is the director of Avatar'', ``entity'': [``Avatar''], ``relation'': [``director'']\} \texttt{</search>} \texttt{<result>} search result here \texttt{</result>} \
\texttt{<think>} This is the reasoning process. \texttt{</think>} \texttt{<answer>} The final answer is \text{\textbackslash boxed\{ answer here \}}\texttt{</answer>}. \
In the last part of the answer, the final exact answer is enclosed within \verb|\boxed{}| with latex format. 
\\
\bottomrule
\bottomrule
\end{tabular}
\caption{System prompt for generating reasoning trajectories through interaction with the environments during training and inference stages.}
\label{system_template}
\end{table}

\begin{table*}[ht]
\centering
\resizebox{1.0\textwidth}{!}{
\begin{threeparttable}[b]
\begin{tabular}{lcccccccccccccc}
\hline
\multirow{2}{*}{Methods} & \multicolumn{3}{c}{\textbf{HotpotQA$^\dagger$}}                               & \multicolumn{3}{c}{\textbf{2Wiki$^\dagger$}}  & \multicolumn{3}{c}{\textbf{Musique$^\dagger$}}   & \multicolumn{3}{c}{\textbf{Bamboogle$^\ddagger$}} & \multicolumn{2}{c}{\textbf{Average}}            \\ \cmidrule(r){2-4} \cmidrule(r){5-7} \cmidrule(r){8-10} \cmidrule(r){11-13} \cmidrule(r){14-15}
                        &F1 & CEM      
                         & EM   & F1  & CEM  & EM & F1& CEM & EM & F1&CEM & EM &F1&CEM\\ 
\hline
\rowcolor[rgb]{0.9,0.9,0.9}
\multicolumn{15}{c}
{\textbf{\textit{Prompt Based}}}  \\
\hline
\multicolumn{15}{l}
{\textbf{\textit{Qwen2.5-7B}}}    \\
Vanilla RAG                 & 29.0    & 22.4        & 20.5     & 32.5     & 27.9& 27.0 & 11.2 &5.1 &3.4 & 17.6 & 12.8 & 10.4 & 22.6 & 17.1     \\
$\text{Iter-RetGen}$      &   51.4    & 45.2    & 39.9      & 39.2     & 35.5       & 32.2  &17.4 &12.4&10.0&31.8&24.8&22.4&34.9&29.5      \\
$\text{IRCoT}$        &   47.2     & 47.3   & 35.3         & 35.0     & 39.2   & 25.5  &14.7 &13.3 &7.5 &32.3 &28.8 &23.2 &32.3 &32.2       \\
$\text{Search-o1}^{*}$       & 36.9  & 32.3     &   27.4        & 41.2     & 41.3   & 33.7  & 16.8 &  13.0& 10.5 &  40.0& 34.4 &30.4 & 33.7&   30.3    \\
  \ \ + \textit{Qwen2.5-32B}     & 56.9  & 51.8   &  44.1   & 64.6  & 68.1  & 55.9 &    28.6& 25.7 &  19.7 & 64.1 &56.8 & 51.2&  53.6 & 50.6   \\
\hline
\multicolumn{15}{l}
{\textbf{\textit{Frontier LLMs}}}    \\

$\text{DeepSeek-R1}$   & 62.5   & 54.0   & 48.0          & 65.7       & 65.0      & 54.0 &39.9&33.0&27.5&  63.0&52.8&52.0&57.8&51.2   \\

$\text{Qwen3-235B-A22B}$        & 57.3       & 56.1       & 44.5       & 59.4        & 64.1         & 45.3 &41.7&39.5&27.6&55.3&49.2&43.8&53.5&52.2          \\
$\text{GPT-4.1}$        & 60.6     & 56.0    & 45.0         & 69.7          & 75.5         & 56.0 &44.9 &47.0 &28.5 &63.8 &55.2 & 49.6 &59.7&58.4       \\
$\text{O4-mini}$        & 57.8   & 59.5    &    40.5      & 62.1  &   71.0    & 47.5 & 41.6 &45.5 &27.5  &61.7 &64.0 &46.4 &55.8 &60.0       \\
$\text{Gemini-2.5-Pro}$        & 55.6&60.5&39.5&71.8&83.0&60.5&37.0&47.0&24.5& 59.7& 69.6& 52.0& 56.0&65.0     \\
\hline
\rowcolor[rgb]{0.9,0.9,0.9}
\multicolumn{15}{c}
{\textbf{\textit{Training Based}}}  \\
\hline
\multicolumn{15}{l}
{\textbf{\textit{Qwen2.5-7B}}}  \\
$\text{Search-R1-v0.3}$     & 61.8 &  53.6  & 49.8      & 60.7      & 58.7     & 52.3   & 30.9 &   24.7  & 21.5   & \textbf{59.4}&   48.0  &\textbf{ 47.2}  & 53.2& 46.3   \\
$\text{ReSearch}$     & 63.2 &  55.8  & 50.4      & 67.1      & 65.4     & 60.3   & 28.0 &   34.1  & 24.0   & 53.1 &   45.6  & 41.6  & 54.4& 48.7   \\
$\text{R1-Searcher}$     & 57.8 &  59.7  & 45.6     & 64.0     & 67.8    & 56.2   & 28.4 &  27.9  & 19.5   & 49.8 &   46.4  & 36.0  & 50.0& 50.5   \\
\rowcolor[rgb]{0.8745, 0.9176, 0.9608}
\multicolumn{1}{l}
{\textbf{DynaSearcher}}          & \textbf{66.1 }& \textbf{62.8} & \textbf{52.0} & \textbf{72.0} &\textbf{76.3}  &\textbf{61.9} & \textbf{38.7}& \textbf{38.6} & \textbf{26.5} & \underline{57.9 }& \textbf{51.2}  & \underline{41.6}  & \textbf{58.7 }&\textbf{57.2} \\

\hline
\multicolumn{15}{l}
{\textbf{\textit{Qwen2.5-32B}}}  \\
$\text{Search-R1-v0.3}$  & 66.5&55.8&53.5&73.4&71.7&68.1&36.2&30.6&28.5&65.1&55.2&54.4&60.3&53.5     \\
$\text{ReSearch}$     & 69.4 &61.0&56.3&78.1&76.7&72.3&39.3&33.8&30.5&63.1&52.0&50.4&62.5&55.9     \\
\rowcolor[rgb]{0.8745, 0.9176, 0.9608}
\multicolumn{1}{l}
{\textbf{DynaSearcher}}        & \textbf{71.3} & \textbf{63.1} & \textbf{57.4} & \textbf{81.9} & \textbf{82.7} & \textbf{75.1} & \textbf{47.4} & \textbf{41.6} & \textbf{35.2} & \textbf{70.3}& \textbf{58.6} & \textbf{55.2} & \textbf{67.7} & \textbf{61.5 }\\ 
\hline
\end{tabular}
\end{threeparttable}}
\caption{Overall evaluation results on the dev or test sets of four benchmarks. The best and second best results are bold and underlined, respectively. All methods are evaluated in the same local retrieval environment. * indicates the results reproduced by us. $^\dagger/\ddagger$ represents in-domain/out-of-domain datasets. + denotes replacement with a new base model.}
\label{primary_table}
\end{table*}

\subsection{Multi-Reward Reinforcement Learning}
Agentic RL with tool integration is an effective training method that endows LLMs with stronger autonomous reasoning capabilities, and has been extensively validated in mathematical and coding scenarios~\cite{feng2025retool,li2025torl}. 
To further enhance the performance of LLMs in open-domain multi-hop question answering tasks, we conduct experiments based on the GRPO RL algorithm~\cite{shao2024deepseekmath}, with the complete rollout process described in Table~\ref{system_template} and Figure~\ref{framework}. To enable the LLM to autonomously interact with the external environments, we introduce three special tokens to standardize the LLM's outputs, \textit{i.e.}, \texttt{<think>}, \texttt{<search>}, and \texttt{<answer>}. During inference, when the LLM generates special tokens such as \texttt{<search>}, it triggers the corresponding search actions to retrieve external knowledge. Through multiple iterations of the \texttt{think}$\rightarrow$\texttt{search}$\rightarrow$\texttt{result} process, the LLM ultimately generates a precise and comprehensive answer within the specific tokens box \texttt{<answer>} and \texttt{</answer>} once it determines that the question has been fully addressed.

During GRPO training, the design of the reward function is a critical component. In \citet{chen2025research,jin2025searchr1,song2025r1searcher}, an outcome-based reward is employed to guide the LLM on learning the reasoning and the capability to call search engine interleaved throughout the RL process. The coarse global reward lacks effective guidance for intermediate query generation and struggles to explore efficient reasoning trajectories. Therefore, we design a multi-reward RL framework to encourage high-quality intermediate query generation, dynamically adapt to different retrieval strategies, and efficiently acquire external knowledge, yielding more stable convergence and higher accuracy in complex reasoning tasks. The reward mechanism mainly includes accuracy, information gain, and penalty rewards.

\subsubsection{Accuracy Reward} 

The accuracy reward includes both format correctness and answer correctness. Following~\cite{shao2024deepseekmath}, we employ an explicit format reward to ensure that the LLM adopts a predefined iterative workflow of ``think$\rightarrow$search'', thereby guaranteeing that the entire agent executes correctly. The required output format is defined in the system prompt, as shown in Table~\ref{system_template}. In addition, for the answer correctness, typical mathematical scenarios use the concrete numerical output as the evaluation criterion. However, this method often fails to provide an accurate and comprehensive evaluation for open-domain QA tasks. We introduce a comprehensive method for evaluating answer accuracy $r_{\text{ans}}$. The complete accuracy reward $r_{\text{acc}}$ is defined as:
\begin{align}
r_{\text{acc}} &=
\begin{cases}
\max(0.1,\, r_{\text{ans}}), & \text{if formate is correct,} \\
0, & \text{if formate is incorrect.}
\end{cases}\\
r_{\text{ans}} &=
\begin{cases}
\text{F}_{1}(a_{\text{pred}},\, a_{\text{gt}}), & \text{if } L_{\text{pred}} \geq n \cdot L_{\text{gt}}, \\
\text{CEM}(a_{\text{pred}},\, a_{\text{gt}}), & \text{if } L_{\text{pred}} < n \cdot L_{\text{gt}}.
\end{cases}
\end{align}
where F1 and CEM are the word-level F1 score and the cover exact match score between the output answer $a_{\text{pred}}$ and the ground truth $a_{\text{gt}}$. $L_{\text{pred}}$ and $L_{\text{gt}}$ denote the number of words in $a_{\text{pred}}$ and $a_{\text{gt}}$. $n$ represents the multiple of the text length, which we set to 3 by default. Under the reward mechanism $r_{\text{ans}}$, our approach strives to generate responses that are both comprehensive and accurate.

\subsubsection{Gain and Penalty Reward} 

Based on the aforementioned accuracy reward, we further introduce an information gain reward $r_{\text{gain}}$ to encourage the generation of high-quality intermediate queries, thus enhancing the performance of document retrieval $r_{\text{recall}}$. Let TP represent the number of relevant documents retrieved, FN for the missed relevant documents, then $r_{\text{recall}}$ can be defined as:
\begin{equation}
r_{\text{recall}} = \frac{\text{TP}}{\text{TP} + \text{FN}}
\end{equation}
where we use the supporting document chunks provided by the dataset as ground truth and then evaluate the relevance of retrieved document chunks based on title matching or embedding similarity scores.

To prevent the LLM from engaging in reward hacking by solely pursuing the $r_{\text{recall}}$, a retrieval penalty reward $r_{\text{penalty}}$ is incorporated into the information gain reward $r_{\text{gain}}$. The $r_{\text{gain}}$ and $r_{\text{penalty}}$ can be calculated as:
\begin{align}
r_{\text{gain}} &= \alpha \cdot (r_{\text{recall}}-r_{\text{penalty}})\\
r_{\text{penalty}} &= \max(\beta, 1 - \gamma^{t-i})
\end{align}
where $\alpha$ is the scale factor used to balance the accuracy reward and the gain reward, $\gamma$ is the decay factor for the penalty reward, and $\beta$ denotes the lower bound of the penalty reward. Here, $t$ denotes the current number of retrieval actions executed by the agent at a rollout, and $i$ represents the ground-truth number of hops or sub-questions for this sample, as annotated in the dataset. When $t$ exceeds $i$, indicating redundant retrievals beyond what is necessary, a penalty is applied to $r_{\text{recall}}$; conversely, if $t$ is less than $i$, a slight reward is granted. This design encourages the LLM to make efficient use of retrieval actions for information acquisition, avoiding excessive and redundant searches. The overall reward can be calculated as:
\begin{equation}
r_{\text{overall}} = r_{\text{outcome}} +  r_{\text{gain}}
\end{equation}
Combining the overall reward with the training objective of GRPO, we propose a reinforcement learning objective that explicitly incorporates a search engine $\mathcal{R}$ during optimization for LLM search agent training~\cite{jin2025searchr1,zheng2025deepresearcher}. The objective is formalized as:
\begin{multline}
\max_{\pi_{\theta}} \mathbb{E}_{x\sim\mathcal{D}, \ y\sim\pi_{\theta}({\cdot|x;\mathcal{R}})} \left[ A_{\phi}(x,y) \right] \\- \beta \mathbb{D}_{\text{KL}}[\pi_\theta(y|x;\mathcal{R}) | \pi_{\text{ref}}(y|x;\mathcal{R})]
\end{multline}
where $\pi_{\theta}$ denotes the trainable policy model, $\pi_{ref}$ is a fixed reference model, $A_{\phi}$ represents the overall advantage function, and $\mathbb{D}_{\text{KL}}$ denotes the KL divergence. Here, $x$ are sampled from the dataset $\mathcal{D}$, and $y$ denote the output sequence interleaving reasoning steps with search engine retrievals. Since the retrieved documents are not generated by the policy model, we mask the retrieval results during loss calculation to prevent the training policy from being biased.

\section{Experiments}
\subsection{Experimental Settings}
\noindent\textbf{Datasets and Evaluation Metrics}. We conduct extensive experiments on six multi-hop datasets, including HotpotQA~\cite{yang2018hotpotqa}, 2WikiMultiHopQA (2Wiki)~\cite{ho20202wiki}, Musique~\cite{trivedi2022musique}, Bamboogle (Bam)~\cite{press2022bam}, MoreHopQA (MoreHQA)~\cite{schnitzler2024morehopqa}, and Frames~\cite{krishna2024frames}. The first three datasets are in-domain datasets, with portions of their training sets used for training, while the latter three are out-of-domain datasets utilized to evaluate the model's generalization performance. Our evaluation is conducted on the full dev or test sets corresponding to the above datasets. For evaluation metrics, we employ the standard word-level F1 score (F1), Cover Exact Match (CEM), and Exact Match (EM). For more complex open-domain QA tasks, we additionally utilize LLM-as-Judge (LasJ) to ensure a fair evaluation.

\begin{table}[ht]
\centering
\setlength{\tabcolsep}{0.31mm}
\resizebox{0.47\textwidth}{!}{
\begin{tabular}{lcccccc}
\hline
\multirow{2}{*}{Methods} & \multicolumn{3}{c}{\textbf{In-Domain}}  &   \multicolumn{3}{c}{\textbf{Out-of-Domain}}                           
\\ \cmidrule(r){2-4} \cmidrule(r){5-7} &  F1   & CEM  & EM     & F1    & CEM &EM        \\ \hline

\textbf{Training} \\
default (doc search + orm)  & 54.2  & 52.3 & 40.9 & 34.4 & 30.3 & 25.0 \\
(a) w/ KG system prompt     & 54.4 & 54.1  & 40.7 & 35.2 & 31.1 & 27.8\\
(b) w/ KG search tool       & 53.9  & 57.1 & 38.0 & 34.3 & 33.6 &  24.7         \\
(c) w/ multi-reward        &  58.9 & 59.2 & 46.8 & 38.8 &34.0 & 30.0       \\
\midrule
\textbf{Inference} \\
(a) w/ Doc search tool only  & 57.1 & 57.6 & 45.2 & 37.3 & 32.4 & 28.8  \\
(b) w/ KG search tool & 58.4 & 58.5 & 46.3 & 38.2 & 33.7 & 29.9 \\ 
(c) w/ Doc+KG filter  & \textbf{58.9} & \textbf{59.2} & \textbf{46.8} & \textbf{38.8} & \textbf{34.0} & \textbf{30.0} \\
\hline
\end{tabular}
}
\caption{Ablation study on various multi-hop datasets (adding one component each time). 'w/' represent 'with'. orm denotes the combination of format and accuracy reward.}
\label{ablation}
\end{table}

\begin{table}[ht]
\centering
\setlength{\tabcolsep}{1.21mm}
\resizebox{0.47\textwidth}{!}{
\begin{tabular}{lccccccc}
\hline
\multirow{2}{*}{Methods} & \multicolumn{2}{c}{\textbf{Bam$^\ddagger$}}  &   \multicolumn{2}{c}{\textbf{Frames$^\ddagger$}}  &   \multicolumn{2}{c}{\textbf{MoreHQA$^\ddagger$}}   &\multicolumn{1}{c}{\textbf{Avg.}}                             
\\ \cmidrule(r){2-3} \cmidrule(r){4-5} \cmidrule(r){6-7} \cmidrule(r){8-8} 
        & F1    & LasJ     & F1    & LasJ     & F1    & LasJ  & LasJ \\ \hline
Search-o1        & 58.6 &  64.6  &24.1 & 31.7   &22.1 & 34.3  &   43.5   \\
ReSearch      & 71.9  & 73.8 & 38.7 & 48.5 & 30.9 & 46.5  &56.3  \\ 
R1-Searcher   & 67.2  & 71.3 & 33.4 & 42.6 & 23.5 & 37.9 &    50.6  \\ \midrule
\textbf{Ours}            & \textbf{75.0} & \textbf{77.8}&\textbf{40.2} &  \textbf{49.2} & \textbf{33.7} &  \textbf{51.8} & \textbf{59.6} \\ \hline
\end{tabular}
}
\caption{Generalization experiments on out-of-domain datasets using online search.}
\label{web_search_table}
\end{table}

\noindent\textbf{Search Tools}.
An efficient search tool is essential for our search agent. We build a local retrieval environment using a dense retriever with the multilingual-e5-base~\cite{wang2022e5} model, incorporating the 2018 Wikipedia corpus~\cite{ho20202wiki}. To obtain more up-to-date and comprehensive information, we further utilize Tavily as a web search tool. Additionally, we build a retrieval service for the knowledge graph based on Wikidata5M~\cite{wang2021wikidata} to obtain triples related to the corresponding entities.
Specifically, we use the Wikidata5M dataset to construct our KG retrieval engine, which contains over 4 million entities and more than 800 relation types. We merge and deduplicate all triples from both the transductive and inductive splits to build a comprehensive entity-relation graph. For each entity name generated by the model during inference, we normalize the case and perform fuzzy matching to identify relevant entities and extract all associated single-hop knowledge subgraphs. These candidates are ranked by the string similarity between the queried entity and relation and those in the KG (i.e., by the number of shared words). Finally, we return the top 100 triples and up to 1024 tokens as the KG retrieval result.

In Wikidata5M, a single entity or relation ID often corresponds to multiple words or phrases, and directly concatenating all of them for input into the LLM can introduce substantial noise. To mitigate this, we introduce a KG filter module. All matched entity–relation pairs, together with the queried entity, relation, current subquery, and the original question, are fed into an LLM. The model then selects and returns up to five of the most relevant entity–relation pairs as the final KG search results.

\noindent\textbf{Baselines and Training Details}.
In our experiments, in addition to comparing with state-of-the-art LLMs such as \textit{DeepSeek-R1-0528}, \textit{Qwen3-235B-A22B}, \textit{GPT-4.1-0414}, o4-mini-0416, and \textit{Gemini-2.5-Pro-0325} (as shown in Table~\ref{primary_table}), we also benchmark against advanced RAG methods~\cite{shao2023iterretgen,trivedi2022ircot,li2025searcho1} and RL-based agentic search models~\cite{jin2025searchr1,chen2025research,song2025r1searcher,zheng2025deepresearcher,wang2025stepsearch}. These experiments are primarily based on the Qwen2.5 models~\cite{qwen2025qwen25technicalreport}, where Qwen2.5-7B and Qwen2.5-32B refer to their respective Instruct models. All training-based models are derived from their corresponding open-source models. The training data of DynaSearcher consist of the stage-2 data from \citet{song2025r1searcher} and 8,000 randomly sampled instances from Musique. The RL training epoch is set to 1, with a train batch size of 16 and a learning rate of 1e-6. The KL divergence coefficient set to 1e-3. Each data sample undergoes 8 rollouts during training. We utilize FSDP~\cite{zhao2023pytorchfsdpexperiencesscaling} and vLLM\footnote{\url{https://github.com/vllm-project/vllm}.} in VeRL\footnote{\url{https://github.com/volcengine/verl}.} framework, with a sampling temperature of 1.0, top-p of 0.95 and a maximum response length of 8192. 
For our information gain and penalty reward, we set $\alpha$ to 0.5, $\gamma$ to 0.9, and $\beta$ to -0.2.

\subsection{Main Results}
Table~\ref{primary_table} comprehensively shows the results of DynaSearcher and other strong baselines on four multi-hop benchmarks.

\begin{figure}[t]
\centering
\includegraphics[width=0.49\textwidth]{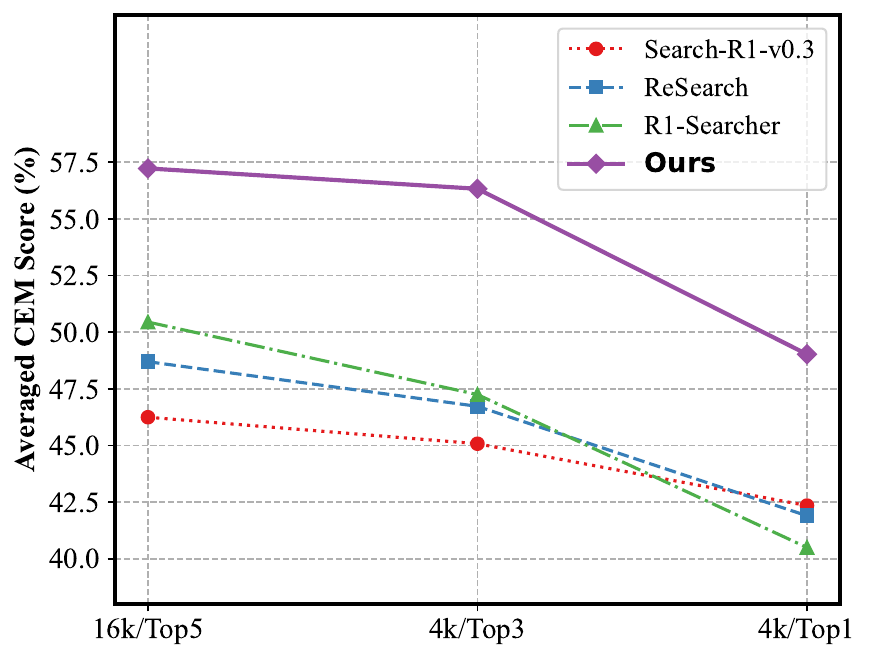} 
\caption{Performance comparison under different inference settings. The averaged CEM score is computed across HotpotQA, 2Wiki, Musique, and Bam. The x-axis denotes the maximum context length and the number of retrieved documents.}
\label{efficiency}
\end{figure}

\noindent\textbf{Achieve significant performance improvements}. Our approach achieves significant performance improvements across multiple benchmarks under all evaluation metrics. These results demonstrate that our approach can effectively guide the reasoning path and achieve a good balance between the comprehensiveness and accuracy of the final responses. Moreover, DynaSearcher still significantly outperforms other baselines under low-resource settings, as shown in Figure~\ref{efficiency}.

\noindent\textbf{Achieve frontier LLM performance with small-scale models}. The current state-of-the-art LLMs are evaluated on various multi-hop datasets. Interestingly, we find that these models struggle to effectively follow the Search-o1 series prompts to guide multi-step reasoning and retrieval. Therefore, we employ a standard RAG approach to achieve better performance. Our DynaSearcher-7B achieves performance on par with GPT-4.1, while surpassing DeepSeek-R1.

\noindent\textbf{Exhibit strong generalization performance}. We further evaluate our approach on larger-scale models and more complex web search environments, where it consistently achieves performance improvements and demonstrates superior performance on out-of-domain datasets. These results show that our approach learns an efficient reasoning paradigm and possesses strong generalization capabilities.

\subsection{Further Analysis}

\subsubsection{Ablation Studies}
To validate the effectiveness of our proposed DynaSearcher framework, we conduct comprehensive ablation studies on key components during both the training and inference stages. The performance of various methods is shown in Table~\ref{ablation}. In the training stage, we design three key variant methods: (a) introduces our designed KG-augmented system prompt based on the default setting; (b) further incorporates our KG search environment on top of (a), enabling access to structured knowledge during training; and (c) further optimizes the original outcome reward on top of (b), providing more fine-grained control over the training objective. In the inference stage, to investigate the relationship between the generalization ability of our approach and the incorporation of additional structured knowledge, we adopt several inference modes: (a) retrieves relevant documents using only the intermediate generated subqueries; (b) further retrieves related subgraphs to guide reasoning process on top of (a); and (c) further filters the retrieved documents and subgraphs on top of (b) to reduce noise during the reasoning process. Experimental results demonstrate that our approach does not merely learn superficial think patterns, but rather acquires the ability to efficiently decompose questions and generate more precise subqueries. This leads to more effective planning strategies and reasoning trajectories.

\subsubsection{Search Environment}

The external information retrieval environment plays a crucial role in search agents. In our training process, we utilize a locally deployed retrieval environment, including both embedding retrieval and KG retrieval (see Table~\ref{ablation}). To further simulate more realistic interactions, we also incorporate online search as an additional evaluation in Table~\ref{web_search_table}. Experimental results show that web search brings significant performance improvements. Meanwhile, during training, the structured information introduced by KG search can effectively guide the model's agentic search process. Even in an environment with only document search, DynaSearcher still demonstrates efficient reasoning capabilities.

\subsubsection{Inference Settings}
Most current search agent methods incorporate the retrieved information into the context, but this method is inevitably constrained by the context length. In Figure~\ref{efficiency}, we compare the performance of our approach and other baselines under low-resource settings. The results show that our DynaSearcher achieves comparable or better performance under the 4k/top1 setting than other methods do under the 16k/top5 setting. This further demonstrates the efficiency and accuracy of the reasoning trajectories enabled by our approach. Therefore, to minimize the context length of the retrieved information, we introduce a document and KG filtering module to eliminate irrelevant content (see Table~\ref{ablation}), which further improves model performance.

\subsection{Qualitative Analysis} 
Table~\ref{case_study} illustrates the iterative reasoning and retrieval process of DynaSearcher. For such complex questions, our model performs a reasonable analysis and decomposition, dynamically adjusting its current strategy based on the retrieved information. For example, in the case presented in the table~\ref{case_study}, relevant temporal and entity information is obtained in the first query. This information is then utilized in the next step to further investigate related entities, thereby reducing unnecessary queries. In addition to full text retrieval, we introduce search term queries, which leverage external tools such as knowledge graphs or web search engines to further improve retrieval effectiveness. 
By integrating these methods, DynaSearcher ultimately achieves both efficient reasoning and accurate answers.

\begin{table}[t]
\centering
    \renewcommand{\arraystretch}{0.8}
\small
\begin{tabular}{p{7.2cm}}
\toprule
\toprule
 \textbf{Question}: What is the name of the play written in May 2016 by a playwright who won the MacArthur Fellowship the same year as the poet who wrote ``Postcolonial Love Poem''? 
 \hfill \textbf{Answer}: \text{Skeleton Crew} \\
\midrule
\texttt{\textbf{DynaSearcher}}: \textbf{\texttt{<think>}} To properly identify the name of the play written in May 2016 by a playwright who won the MacArthur Fellowship the same year as the poet who wrote ``Postcolonial Love Poem'', I first need to determine who that poet is and then find out who the playwright who won the MacArthur Fellowship in the same year is and what play they wrote in May 2016. \textbf{\texttt{</think>}}

\textbf{\texttt{<search>}} \{``query'': ``which poet wrote the book `Postcolonial Love Poem'?'' , ``entity'': [``Postcolonial Love Poem''], ``relation'': [``author name'']\} \textbf{\texttt{</search>}}

\textbf{\texttt{<result>}} ... Natalie Diaz is the author of Postcolonial Love Poem. In 2018, she received a MacArthur Fellowship ... \textbf{\texttt{</result>}}\\
\textbf{\texttt{<think>}} ... \textbf{\texttt{</think>}}\\

\textbf{\texttt{<search>}} \{``query'': ``who is the playwright that won the MacArthur Fellowship in 2018?'', ``entity'': [``2018 MacArthur Fellowship''], ``relation'': [``recipient name'']\} \textbf{\texttt{</search>}}\\

\textbf{\texttt{<result>}} ... \textbf{\texttt{</result>}}\\
\textbf{\texttt{<think>}} ... \textbf{\texttt{</think>}}\\

\textbf{\texttt{<search>}} \{``query'': ``what is the name of the play written by Dominique Morisseau in May 2016?'', ``entity'': [``Dominique Morisseau''], ``relation'': [``name of play written in May 2016'']\} \textbf{\texttt{</search>}}\\

\textbf{\texttt{<result>}} ... \textbf{\texttt{</result>}}\\
\textbf{\texttt{<think>}} ... \textbf{\texttt{</think>}}\\

\textbf{\texttt{<answer>}} The final answer is \text{\textbackslash boxed\{Skeleton Crew\}} \textbf{\texttt{</answer>}}\\ 

\bottomrule
\bottomrule
\end{tabular}
\caption{Quantitative analysis of the efficient reasoning process in Frames dataset.}
\label{case_study}
\end{table}

\section{Conclusions}
In this paper, we propose DynaSearcher, a dynamic knowledge graph augmented multi-reward reinforcement learning framework tailed for search agents, enabling flexible interaction between the reasoning process and the external environment. The incorporation of structured knowledge efficiently guides the search process by modeling entity relationships, while multi-reward RL enables nuanced control over training objectives. Extensive experiments on six complex benchmarks demonstrate the superiority and efficiency of our approach, achieving performance on par with current frontier LLMs using only a small-scale model and limited computational resources. Furthermore, we validate the generalization and robustness of our approach across diverse retrieval environments and larger-scale models, highlighting its substantial potential for broad applicability.

\bibliography{custom}

\clearpage
\appendix
\section{Implementation Details}
\begin{figure*}[t]
\centering
\includegraphics[width=1.0\textwidth]{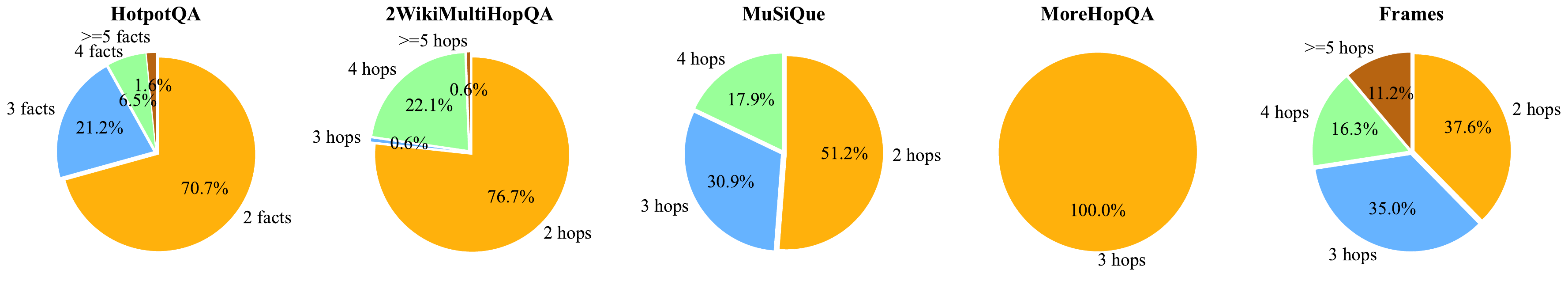} 
\caption{Overview of the distribution of query complexity over five multi-hop QA datasets.}
\label{hop_dist}
\end{figure*}

Due to the large size of the dev sets in the 2WikiMultiHopQA and HotpotQA datasets, which affects iteration efficiency, we randomly sample 1,000 examples from their respective dev sets as our final test set, with a fixed random seed 42. We also verify that the performance on this subset is nearly identical to that on the full dev set, indicating that this approach can significantly improve iteration efficiency. To
better understand the complexity of multi-hop reasoning in these datasets, we analyze the hop distribution of the HotpotQA, 2WikiMultiHopQA, MuSiQue, MoreHopQA, and Frames dev/test sets in Figure~\ref{hop_dist}. The statistics show that there is a high proportion of complex reasoning queries with 3 hops or more. HotpotQA lacks explicit hop annotations, so we instead count the number of supporting facts.

In the retrieval process, we employ the \textit{multilingual-e5-base} as the
retriever and use the widely used Wikipedia
dump from December 2018 as the retrieval corpus, which comprises over 21 million passages. To improve retrieval efficiency, we combine the supporting document passages from five multi-hop datasets with one million randomly sampled documents from the 2018 Wikipedia dump to create our final retrieval corpus.

During the training phase, our training data consist of a total of 8,148 examples from HotpotQA and 2WikiMultiHopQA, which are selected through data selection in R1-Searcher. In addition, we randomly sample 8,000 examples from the training set of MuSiQue to form our final training set. The training consists of 2 epochs, with a \verb|train_batch_size| of 16 and a learning rate of 1e-6. \verb|ppo_mini_batch_size| is set to 16. The maximum lengths for prompt and response are set to 512 and 8192. Rollouts are conducted with a batch size of 8 and a temperature of 1.0 to encourage exploration. The KL-divergence regularization coefficient and the clipping ratio are set to 0.001 and 0.2, respectively. All experiments are carried out on eight NVIDIA-H20-96G. In the inference stage, we use SGLang or vLLM as the underlying inference engines and set different maximum context lengths and maximum retrieval times to avoid the impact of outlier samples on training. For the evaluation of other prompt-based baselines, we use the implementations provided in the ReSearch GitHub repository\footnote{\url{https://github.com/Agent-RL/ReCall}.}. For other training-based methods, we evaluate them using their publicly available trained models.

\section{Prompt Examples}
Although the retrieval process of the knowledge graph (KG) can provide more diverse information for question exploration, it can also introduce noise for the analysis of the initial question. Therefore, we introduce a KG filtering module to control the number of retrieved triples. The specific filtering logic is shown in Table~\ref{kg_extract_prompt}.
Table~\ref{judge_prompt} presents the implementation prompt for our LLM-as-Judge (LasJ) score. By leveraging larger and more powerful LLMs as judge models, we can achieve more accurate judgments of responses in the multi-hop QA scenarios. This more precise evaluation approach can be incorporated into the training process, which also introduces additional computational overhead for training.

\begin{table}[ht]
\centering
    \renewcommand{\arraystretch}{0.8}
\small
\begin{tabular}{p{7.2cm}}
\toprule
\toprule
You are an information extraction expert. Given an input of query, entities and relations, 
your task is to extract the most relevant evidence(s). If there is no relevant one, return "None".
The names of entities or relations can be multiple alias, retain the exact one.

For example, \\
\texttt{<input>}
\{``query'': ``Who is the director of Avatar'', ``entity'': [``Avatar''], ``relation'': [``director'']\}
\texttt{</input>}

\texttt{<result>} \\
\{``entity1\textunderscore{}id": ``x1", 
         ``entity1": ``Avatar", 
         ``relation\textunderscore{}id": ``y1", 
        ``relation": ``distinct element of; has class", 
        ``entity2\textunderscore{}id": ``x2", 
         ``entity2": ``character code; codeset"\} 
        
\{``entity1\textunderscore{}id": ``x1",                 ``entity1": ``Avatar",                    ``relation\textunderscore{}id":                 ``y2", 
        ``relation": ``director;             directed", 
        ``entity2\textunderscore{}id": ``x3",  ``entity2": ``James Cameron; james cameron; James' Cameron"\}

\texttt{</result>}

start extracting: \\
\texttt{<evidence>}
entity1; relation; entity2
Avatar; director; James Cameron
\texttt{</evidence>}

Now, start!
\\
\bottomrule
\bottomrule
\end{tabular}
\caption{Prompt for kg filtering module in retrieved triples.}
\label{kg_extract_prompt}
\end{table}

\section{Quantitative Analysis}
Table 1 in the main content presents the performance of state-of-the-art LLMs on multi-hop question answering tasks. Interestingly, we find that these models struggle to effectively follow instructions under the search-o1 paradigm, resulting in suboptimal performance. Additionally, in the basic RAG setting, where models are simply asked to answer questions based on retrieved documents, the models tend to respond that no relevant information is found when the answer is not present in the retrieved documents, failing to fully utilize their inherent capabilities. Therefore, after optimizing the prompt for the RAG scenario (see Table~\ref{rag_plus_prompt}), the models are able to better integrate their own fundamental abilities with the retrieved information to jointly solve the original questions. As a result, Table 1 in the main text content demonstrates the superior performance of frontier LLMs.

\begin{table}[ht]
\centering
    \renewcommand{\arraystretch}{0.8}
\small
\begin{tabular}{p{7.2cm}}
\toprule
\toprule
You will be provided with three pieces of content: the questioner's question, the user's response, and the reference answer list.
Your task is to score the accuracy of the user's response based on the criteria outlined below.
Please ensure that you carefully read and understand these instructions.\\
Evaluation Criteria:\\
Accuracy - Whether the user's answer is consistent with the reference answer and answers the questioner's question. We define this dimension as "whether the user's response includes all the key points from the reference answer and answers the questioner's question."\\
Evaluation Steps:\\
1. Carefully read the questioner's question and understand its key points.\\
2. Carefully read the reference answer and understand the key points relevant to the question.\\
3. Check whether the user's response includes all the key points from the reference answer and answers the questioner's question.\\
4. Based on the evaluation criteria, assign a score in the range of 0 to 5, where 0 indicates that the user's response does not include any of the key points from the reference answer and completely fails to answer the questioner's question; 5 indicates that the user's response includes all the key points from the reference answer and fully and correctly answers the questioner's question.\\
Example:\\
Questioner's question:\\
\{question\}\\
Reference answer:\\
\{answer\}\\
User's response:\\
\{response\}

Evaluation result (output only the score between 0 and 5):
\\
\bottomrule
\bottomrule
\end{tabular}
\caption{Judge prompt for LLM-as-judge scoring.}
\label{judge_prompt}
\end{table}

\newpage

\begin{table}[ht]
\centering
    \renewcommand{\arraystretch}{0.8}
\small
\begin{tabular}{p{7.2cm}}
\toprule
\toprule
You are an expert in question answering. Given a question within \texttt{<question>} \texttt{</question>} \
and some contexts within \texttt{<context>} \texttt{</context>}, you first think about the reasoning process within \texttt{<think>} \texttt{</think>} \
and put the answer within \texttt{<answer>} \texttt{</answer>}. \\
For example, \texttt{<question>} This is a question \texttt{<question>} \
\texttt{<context>} Here are contexts \texttt{<context>} \texttt{<think>} This is the reasoning process. \texttt{</think>} \
\texttt{<answer>} The final answer is \text{\textbackslash boxed\{ answer here \}}\texttt{</answer>}. \
If the answer could not be deduced from the contexts or it's wrong, give the right answer based on your own knowledge. \
If the question is ambiguous or the contexts contain multiple possible answers, \
list all possible answers within \verb|\boxed{}| with latex format, separated by commas.
\\
\bottomrule
\bottomrule
\end{tabular}
\caption{Prompt for vanilla retrieval augmented generation.}
\label{rag_plus_prompt}
\end{table}

\end{document}